# Machine Learning and Thermography Applied to the Detection and Classification of Cracks in Building

Angela Busheska[1], Nara Almeida[2], Nicholas Sabella[1], Eudes de A. Rocha[3]

[1] Lafayette College, [2] University of Washington - Tacoma, [3] University de Pernambuco

*Abstract*— Due to the environmental impacts caused by the construction industry, repurposing existing buildings and making them more energy-efficient has become a high-priority issue. However, a legitimate concern of land developers is associated with the buildings' state of conservation. For that reason, infrared thermography has been used as a powerful tool to characterize these buildings' state of conservation by detecting pathologies, such as cracks and humidity. Thermal cameras detect the radiation emitted by any material and translate it into temperature-color-coded images. Abnormal temperature changes may indicate the presence of pathologies, however, reading thermal images might not be quite simple. This research project aims to combine infrared thermography and machine learning (ML) to help stakeholders determine the viability of reusing existing buildings by identifying their pathologies and defects more efficiently and accurately. In this particular phase of this research project, we've used an image classification machine learning model of Convolutional Neural Networks (DCNN) to differentiate three levels of cracks in one particular building. The model's accuracy was compared between the MSX and thermal images acquired from two distinct thermal cameras and fused images (formed through multisource information) to test the influence of the input data and network on the detection results.

## I. Introduction

With the onset of climate change and the subsequent rise in global temperatures, countries worldwide have sharpened their focus on energy efficiency and emissions reduction. The burden of developing solutions to the emerging crisis falls on the government, industry, and academia. While governments establish laws, parameters, and guidelines that encourage or oblige the industry to adopt sustainable practices, academia re-examines these practices through the lens of the economy, society, and the environment; and develops novel methods and ideas, striving to mitigate the climate crisis.

Buildings represent an enormous, untapped reservoir of potential energy use savings and greenhouse gas emissions reductions. Building and construction industry globally represents about 33% of global energy consumption and almost 25% of carbon dioxide emissions [1]. If the whole lifecycle of a building is considered, buildings would, directly and indirectly, account for about 37% of the global $CO_2$ emissions related to energy [2]. In the US, buildings represented 34% of the country's total energy consumption in 2021 [2], and 24% of the American construction and demolition waste was still sent to landfills in 2018 [3]. These significant impacts on the built environment are partially justified by buildings' demand for resources and energy in every phase of their life cycle.

Throughout their lifespan, buildings experience several fates, such as being (1) abandoned in their original condition and deteriorating; (2) reused without major adaptive modifications; (3) significantly modified and repurposed; or (4) demolished. Of these approaches, the first is the worst-case scenario [4], resulting in a wastage of land, material, and energy, jeopardizing the pillars of sustainability. On the other hand, the most sustainable solution tends to be the adaptive reuse of the building, and if this strategy is adopted, the use of thermography may gain relevance in the diagnosis of pathologies, such as cracks, or presence of humidity.

There are a significant number of variables that may affect the inputs and outputs of thermographic tests in buildings, making its interpretation challenging. Thus, technologies have been used to support the automation of pathology detection, which includes the use of sensors, multi-vision cameras, and 3D laser scanning. Among these, the ones that are digital image-based have been applied more widely, due to its low cost, fast-processing speed, and high robustness. Chen et al [5] state that, if compared to multi-visual cameras or laser-scanned points or images, thermal imagers usually present better real-time efficiency, besides lower costs, and the ability to process raw data directly into deep learning networks. However, complex pathologies may not be detected through the use of two-dimensional images only [5], and therefore other technologies, such as machine learning, may come in handy in the fields of architectural heritage preservation.

Gopalakrishnan [6] notes the relevance of deep machine learning - a subset of machine learning that recognizes patterns, mimicking the neural network of human brains - in digital image-based pathologies detection. Deep learning is a novel method that describes a family of learning algorithms rather than a single process, and it can be used to learn complex prediction models, because it represents a multi-layer neural network with many hidden units. Deep learning models originate from a software system that operates in a similar way to human neurons [7], and every layer decides its input for itself and then sends this information to the next tier of nodes.

The main advantage of using Deep Learning in digital image-based pathology detection is that it increases the accuracy of the model and the reliance on it [5][8]. However, it is important to note that Deep Learning is still at its best when rough-ready results are the desired output instead of perfect results [9], which makes its use more appropriate in the initial qualitative analysis of buildings. Therefore, Deep Learning techniques applied to thermography would most likely succeed in the reconnaissance or initial levels of detail of pathology documentation [10][11].

According to Chen et al. [5], methods such as SVM (Support Vector Machine), RBF (Radial Basis Function), KNN (K-Nearest Neighbor), and Random Decision Forest might also be used to identify some pathologies. Several authors have utilized deep learning and artificial intelligence methods to identify pathologies in buildings more efficiently [12][5][6]. A specific deep learning method, called Image classification, reaches a human-like knowledge recognizing specific elements graphically represented in pictures. This image classification method has therefore a great potential in the identification and classification of building defects and pathologies, and it was one of the technologies utilized in this research project.

## II. METHODS

Given the increasing need for innovative technologies and sustainable practices in the construction industry, this research paper aims to develop an image recognition and machine learning model of high accuracy, and further design an application that identifies specific pathologies in buildings using infrared thermography. The focus was given to three different levels of cracks, which are defined by the change in temperature (ΔT) between the crack and its surroundings. The first level types of cracks are those in which ΔT is less than 2°C, while the second level includes those in which ΔT is between 2°C and 4°C, and the most concerning level of cracks denote a ΔT greater than 4°C.

All the data was acquired in Easton, Pennsylvania (marked as the red dots in Figure 2). Two buildings, the McKelvy House and the Laundry Building were used extensively to build the data set due to their structure and the presence of cracks. The buildings were chosen due to their age and structure which contained cracks.

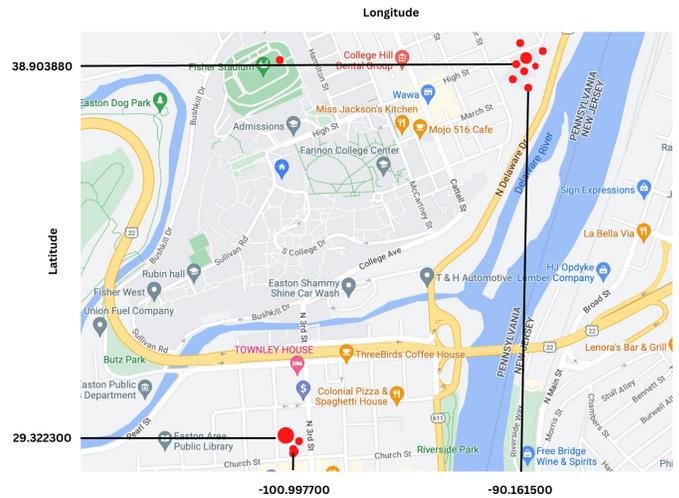

**Figure 1: Data Collection Area**

First, infrared thermographic tests were performed on cracks of random buildings located in Easton, Pennsylvania from June to August of 2022, always right after sunrise time. During these tests, we used HTI HT-18 (220 x 160 pixels) and FLIR One Pro (160 x 120 pixels) cameras to take pictures of each specific crack. The idea was to have MSX images, obtained from the FLIR One Pro, and conventional and thermal images, taken with the HTI HT-18, so all these images could be included in the ML model, and their accuracy levels could be tested and compared.

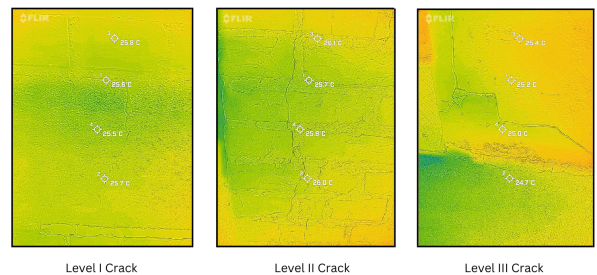

Level I Crack    Level II Crack    Level III Crack

**Figure 2: Data Examples of FLIR MSX Pictures**

All the FLIR MSX thermal images were evaluated using the camera's software, so the three different levels of cracks could be identified. Further image treatments were performed on some FLIR MSX images for addition and reduction of noises, improved quality, or enhanced accuracy of the model, using MatLab.

Also utilizing MatLab, conventional and thermal images captured using the HTI HT-18 camera were fusioned, by applying a specific function that used a 50% transparency filter in the thermal images, superposing it on top of the conventional
images of the same crack. To avoid any potential distraction of training the model, all pictures were resized to 1080 x 1440 pixels with three channels for RGB (Red, Green and Blue). To ensure the highest level of clarity, on selected pictures it was necessary to remove noise and further sharpen the image. Such noise includes a blurry image or unnecessary objects present in the background.

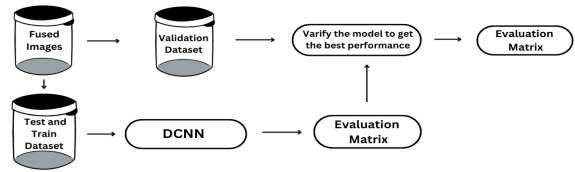

Figure 5: Workflow of Fused Pictures

The model used to evaluate the pictures is DCNN. This convolutional neural network model was proposed by the University of Oxford. It consists of a neural network of layers. It is a classification method that can recognize the different colors of the spectrum.

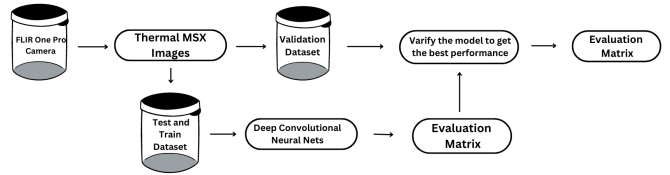

Figure 6: Workflow of FLIR MSX Pictures

To make sure that the model is well trained to focus on essential items, we have extracted two important features: the shape of the crack, and the respective temperature difference. That helps the model to detect if a crack is present in a certain picture, and then classify it accordingly.

The project uses eleven layers of neural network with six different layers. The function of each layers is explained in the figure 7.

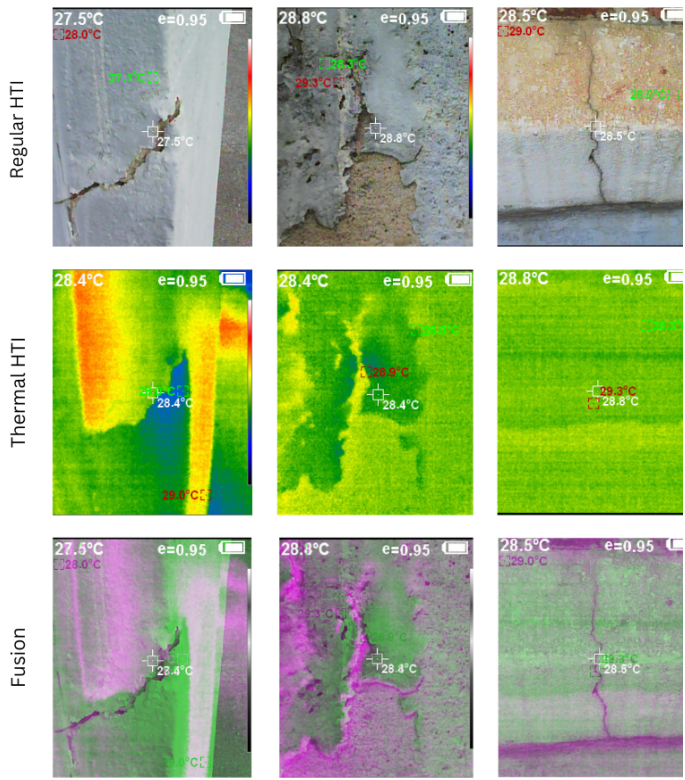

Figure 3: Data Examples of HTI - Fusion Pictures

During image processing, we utilized the previously mentioned method of deep learning (DL), called Image classification. For that, 80% of the images were used in the training set, and 20% for testing and validating.

| No.of Layers | Description | Details |
|---|---|---|
| 1 | Image Input | Receiving images from the previously created datastore |
| 2 | Convolution | Extracts features from images for identification |
| 3 | Max Pooling | Reduces size of the training data produced by the convolution layer |
| 4 | Convolution | Extracts features from images for identification |
| 5 | Max Pooling | Reduces size of the training data produced by the convolution layer |
| 6 | Convolution | Extracts features from images for identification |
| 7 | Max Pooling | Reduces size of the training data produced by the convolution layer |
| 8 | Flatten | Collapses the spatial dimensions of the input into the channel dimensions |
| 9 | Dense | Aligns extracted features to certain features to aid in future testing |
| 10 | Dense | Aligns extracted features to certain features to aid in future testing |
| 11 | Output | Classifies the given data into the correct categories |

Figure 7: Function of the Convolutional Layers

The selected learning rate was initially chosen to be 0.1. When we've selected a higher one, a lot of details were not reflected. On the contrary, if the learning rate was really low, then it would take a lot time to evaluate the dataset.

III. RESULTS

The results of our study are evaluated using metrics like accuracy, precision, recall, and F1 score based on their respective true positive (TP), false positive (FP), true negative (TN), and false negative (FN) values.

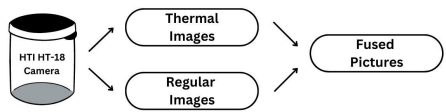

Figure 4: Combination of Thermal and Regular Images

These metrics were computed as follows:

$$Accuracy = \frac{TP + FN}{TP + TN + FP + FN} \quad (1)$$

$$Precision = \frac{TP}{TP + FP} \quad (2)$$

$$Recall = \frac{TP}{TP + FN} \quad (3)$$

$$F_{score} = \frac{Precision * Recall}{Precision + Recall} \quad (4)$$

The datasets used in this study were randomly divided into training, validation, and test sets, accounting for 60%, 20% 20% split respectively. Figure 6 presents the original and the fusion dataset size.

| Data Set | Total | Train (60%) | Validation (20%) | Test (20%) |
|---|---|---|---|---|
| FLIR MSX Pictures | 1355 | 813 | 270 | 270 |
| HTI - Fusion Pictures | 1550 | 900 | 300 | 300 |

**Figure 8: Data Set & Split Information**

Although other models were explored in this research, the ones selected showed the most promising results. The main task the machine learning model was to recognize the three different types of cracks. After the first several epochs, we were able to receive good accuracy. The achieved accuracy was higher than 96%. The greatest limitation in the process was the recognition of the first-level cracks, as some couldn't be distinguished from the structure of the wall.

**Figure 9: Evaluation matrix of the proposed detection method**

The results of the model are given in table 10. The results show that the FLIR MSX Pictures have higher accuracy, precision, recall, and F1 score.

| Image Type | Accuracy | Precision | Recall | F1 |
|---|---|---|---|---|
| FLIR MSX Pictures | 96.83% | 96.91% | 96.73% | 96.83% |
| HTI - Fusion Pictures | 96.12% | 95.65% | 95.77% | 96.12% |

**Figure 10: Evaluation matrix of the proposed detection method**

IV. CONCLUSION

The preservation of architectural heritage and its adaptive reuse are, more than ever, contemporary, and relevant issues that have been taken seriously in several countries. The use of thermography has become more common in this field of application, as it facilitates the detection of defects and pathologies in buildings. However, several factors must be considered in the thermographic analysis, and it is easy to commit mistakes and misinterpretations while reading thermal images. This research project aims to promote an easy reading of thermal images by combining it with machine learning and artificial intelligence, so the diagnosis of building pathologies may be accessible to all.

As governments and local authorities are embracing adaptive reuse, this research project offers the basis for a valuable analysis and classification tool utilizing thermal imagery. In order to speed up thermal mapping and increase location accessibility for the use of this program, one future path would be to integrate this research with thermal imaging drones. Faster, smaller, and with more accessibility than humans, drones mounted with thermal cameras could feed imagery data to this machine learning model for classification and analysis. This method would provide a quicker, more practical, and cost-effective diagnosis of target buildings. Such a scenario is not unprecedented, given the future development path of smart cities and the Internet of Things. As such, this research project which aims to promote the use of machine learning and AI applied to thermal imagery analysis does have practical, real-world applications which could be subjected to further development and refinement.